\documentclass[letterpaper, 10 pt, conference]{ieeeconf}  

\IEEEoverridecommandlockouts                              

\usepackage{multicol}
\usepackage[bookmarks=true]{hyperref}
\usepackage{multirow}
\usepackage{times}
\usepackage{amsmath,amssymb,amsfonts}
\usepackage{algorithmic}
\usepackage{graphicx}
\usepackage{textcomp}
\usepackage{xcolor}
\usepackage{siunitx}

\title{\LARGE \bf
An Efficient Metric for Data Quality Measurement in Imitation Learning 
}

\author{Noushad Sojib$^{1}$, Sajay Arthanat$^{2}$, and Momotaz Begum$^{1}$
\thanks{$^{1}$Noushad Sojib and Momotaz Begum are with the Department of Computer Science, University of New Hampshire, USA {\tt\small noushad.sojib@unh.edu}, {\tt\small mbegum@cs.unh.edu}}%
\thanks{$^{2}$Sajay Arthanat is with the Department of Occupational Therapy and the Center for Digital Health Innovation, University of New Hampshire, USA {\tt\small sajay.arthanat@unh.edu}}%
}

\begin{document}

\maketitle
\thispagestyle{empty}
\pagestyle{empty}

\begin{abstract}
Imitation learning (IL) has seen remarkable progress, yet field deployment of IL-powered robots remains hindered by the challenge of out-of-distribution (OOD) scenarios. Fine-tuning pre-trained policies with end-user demonstrations collected in deployment environments is a promising strategy to address this challenge. However, end-user demonstrations are frequently of poor quality, characterized by excessive corrective motions, oscillations, and abrupt adjustments that degrade both learned and fine-tuned policy performance. Existing automated approaches for curating demonstration data require policy rollouts in the environment, making them computationally expensive and impractical for real-world deployment. In this paper, we propose a fast, efficient, and fully automated demonstration ranking metric based on the power spectral density (PSD) of demonstration trajectories. The PSD metric requires no policy learning, environment interaction, or expert labeling, making it well-suited for scalable, in-the-field data curation. Lower PSD values correspond to smoother, higher-quality demonstrations, while higher PSD values indicate erratic, artifact-laden trajectories. We evaluate the proposed metric on two benchmark imitation learning datasets comprising expert and lay-user demonstrations, and through a user study with older adults at a retirement facility, where collected demonstrations are used to fine-tune $\pi0.5$  \cite{intelligence2025pi_} for a daily living task. Results demonstrate that PSD-curated data yields policies with higher task success rates and smoother execution trajectories compared to uncurated baselines and two competitive data-ranking methods.

\end{abstract}

\section{INTRODUCTION}
Imitation Learning is experiencing its heyday, catalyzed by incredible progress in large data collection \cite{open_x_embodiment_rt_x_2023,team2024octo}, compact visual representation learning \cite{intelligence2025pi_, bjorck2025gr00t}, sim2real \cite{pan20251001demos,yang2025novel}, and end-to-end policy learning \cite{chi2023diffusionpolicy,zhang2024affordance, zhao2023learning}. There has been no better time to think about deployment of IL-powered robots in homes to benefit older adults or people with disabilities in their activities of daily living \cite{sojib2024self}. One of the primary impediments toward field deployment of IL policies is their inability to deal with out of distribution (OOD) scenarios which are unavoidable at run-time, especially in natural home environments. Recent progress in IL policy learning has primarily been driven by high-quality expert demonstrations, making the resulting policies brittle at run-time due to OOD occurrences \cite{intelligence2025pi_, team2024octo}. Fine-tuning of pre-trained policies with end-users' demonstration data collected from the deployment settings is a natural choice to combat OOD errors and make IL usable in the field \cite{ahn2022can}. Apart from fine-tuning, the ability to learn new policies directly from end-users data is extremely important for field deployment since pre-trained policy may not be available for every task that an end-user wants the robot to do for him/her. Demonstrations from end-users not only increases the training data size but also introduces valuable learning signals arising from diverse human strategies and execution styles.

End-user demonstrations are often inefficient due to unstable control and inconsistent task execution \cite{robomimic2021}. Novice teleoperation, for instance, tends to produce excessive corrective motions, oscillations, and abrupt adjustments that deviate from task intent. Kinesthetic teaching of high-DOF robots by lay users introduces similar artifacts. Critically, these artifacts persist across tasks even when task completion is nominally achieved. Training a policy entirely on such low-quality data yields limited success rates. Moreover, when used for fine-tuning, even a small proportion of poor-quality demonstrations can disproportionately degrade a pre-trained policy — in both task success rate and trajectory quality. 
A standard mitigation strategy is to filter poor-quality demonstrations from the training pool prior to policy training or fine-tuning. Manual and semi-manual filtering approaches are tractable at small dataset scales but become impractical as data volume grows \cite{khazatsky2024droid}. Most of the existing automated curation methods require policy rollouts in the environment to rank demonstrations \cite{agia2025cupid, chen2025curating}, or expensive latent space calculation \cite{hejna2025robot} which entails many hours of computation and supervision. These pipelines demand significant robotics expertise and are ill-suited for practical deployment. \textbf{What is needed instead is a fast, reliable, and automated method for ranking demonstrations prior to policy learning or fine-tuning — one that does not depend on environment interaction or expert oversight}. To address this, we propose an efficient, automated metric for ranking demonstrations that requires neither policy learning nor environment interaction, nor any form of expert labeling. The metric is based on the power spectral density (PSD) of the demonstration trajectory: lower PSD corresponds to smoother, higher-quality demonstrations characterized by greater task success and fewer motion artifacts, while higher PSD indicates inefficient demonstrations marked by erratic motion and frequent task failure.

We evaluate the proposed PSD metric by assessing its impact on downstream policy performance, across both policy learning from scratch and fine-tuning settings. We apply the metric to curate two benchmark imitation learning datasets comprising demonstrations from both expert and lay users \cite{robomimic2021, sojib2024self}, and evaluate the performance of policies trained on the curated data. Additionally, we conduct a user study with older adults at a retirement facility to collect lay-user demonstrations for fine-tuning the state of the art pre-trained VLA model $\pi$ 0.5 \cite{intelligence2025pi_} on a simple daily living task. Results demonstrate that data curated using the proposed PSD metric yields policies with higher task success rates and smoother execution trajectories compared to uncurated baselines and two competitive data-ranking methods \cite{agia2025cupid,hejna2025robot}.     
Our contributions are as follows:

\begin{itemize}
    \item \textbf{A training-free demonstration ranking metric}: We propose using power spectral density (PSD) of demonstration trajectories as an efficient, automated metric for ranking demonstration quality, requiring no policy learning, environment interaction, or expert labeling.
    
    \item \textbf{Validation across learning and fine-tuning settings}: We demonstrate the PSD metric's effectiveness on two benchmark IL datasets and a real-world user study with older adults, showing that PSD-curated data consistently yields higher task success rates and smoother trajectories compared to uncurated baselines and competitive ranking methods.
\end{itemize}

\begin{figure*}
\centering
\includegraphics[width=6.1in, height=3.6in]{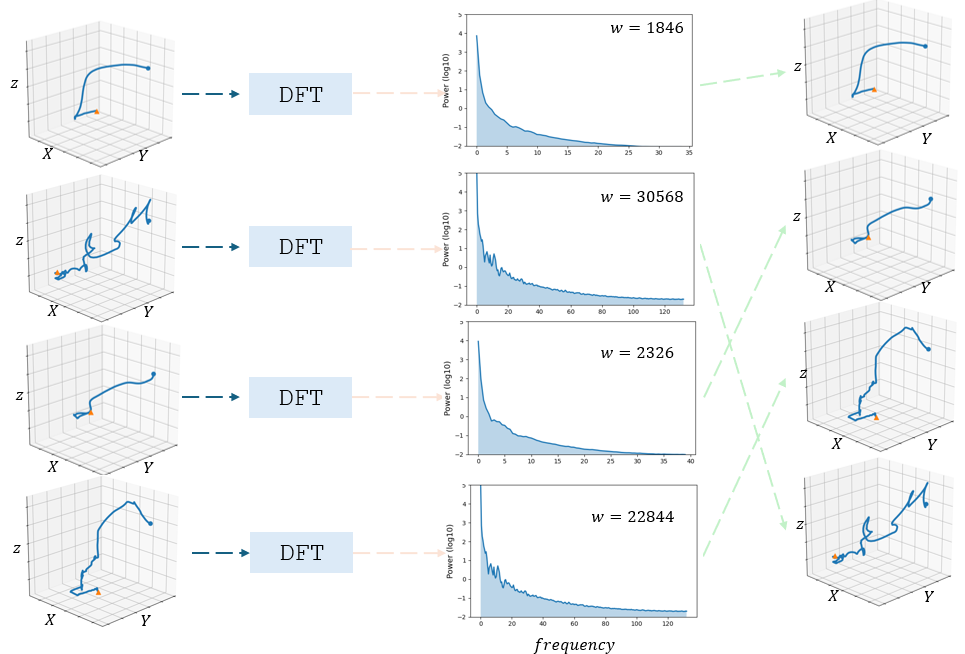}
\caption{Demonstration trajectories (left) and their power spectrum (middle). Poor demonstrations exhibit broader spectral support and higher total power while good demonstrations are strong in low frequencies. Demonstrations ranked according to power spectral density are shown in the right column.
}
\label{fig:concept}
\end{figure*}

\section{Related Work}
A growing body of work studies imitation learning under suboptimal or mixed-quality demonstrations across settings such as offline imitation learning, reinforcement learning, and inverse reinforcement learning. In this section, we focus on prior work most relevant to offline imitation learning. We organize the literature into three categories: (i) data quality, which examines what aspects of demonstrations affect policy performance; (ii) policy learning from suboptimal data, which directly trains policies on mixtures of expert and non-expert demonstrations; and (iii) learning to rank, which aims to identify and order demonstrations by quality, either at the trajectory or segment level. Our work falls into the third category.

\subsection{Data Quality}
Several works highlight the importance of demonstration quality in offline imitation learning. RoboMimic \cite{mandlekar2021matters} shows that removing the lowest-quality demonstrations from a multi-human, mixed-quality dataset can significantly improve policy performance. \cite{belkhale2023data} further demonstrates that naively selecting diverse data does not always yield better results, emphasizing that diversity alone is insufficient without accounting for quality. \cite{sakr2025consistency} identify trajectory path length and jerk as indicators correlated with policy success; however, their evaluation is limited to relatively simple tasks.

\subsection{Policy learning from suboptimal data}
Another line of work focuses on learning policies directly from suboptimal or noisy demonstrations. BCND \cite{sasaki2020behavioral} iteratively learns a policy using an ensemble-based approach, but relies on the assumption that the majority of demonstrations are of high quality. ILEED \cite{beliaev2022imitation} jointly estimates latent expertise labels while learning the policy, enabling robustness to mixed-quality data. DP-IL \cite{wang2023imitation} first purify imperfect demonstrations using diffusion process and then learn policy on top of the purified data. This process is computationally expensive for high dimensional states such as images.

\subsection{Learning to rank demonstrations}
Several prior works propose scoring mechanisms to identify high-quality demonstrations within mixed-quality datasets. CUPID\cite{agia2025cupid} leverages influence functions to estimate the contribution of individual demonstrations to downstream policy performance. This approach requires executing a learned policy in the environment to estimate influence which can be risky for real-world robots, particularly when policies are trained on mixed-quality data. Moreover, influence estimation becomes expensive for real robot complex tasks due to the large number of rollouts required. Demonstration-Information (DemInf) \cite{hejna2025robot} estimates demonstration quality using mutual information, relying on $k$-NN–based estimators. However, empirical studies report weak correlation between DemInf rankings and downstream policy performance \cite{agia2025cupid, zhang2025scizor}. Demo-Score \cite{chen2025curating} similarly depends on executing an initially trained policy to collect rollouts that are subsequently labeled as successful or failed. This reliance on online interaction restricts its applicability to real-world robotic systems where large-scale rollouts are costly or unsafe. DataMIL \cite{dass2025datamil} models the relationship between data and policy performance and proposes an offline metric for demonstration quality without requiring environment interaction. BED \cite{sojib2024self} adopts a self-supervised approach that selects high-quality demonstrations using multi-term consistency metrics. However, BED assumes that the majority of demonstrations are of high quality, an assumption that may not hold for datasets collected from non-expert users. S2I \cite{chen2025towards} applies contrastive learning to classify extracted trajectories as good or bad. While effective at separating trajectory clusters, its preference-learning variant requires a fixed set of high-quality demonstrations a priori, limiting its practicality in fully uncurated datasets. SCIZOR \cite{zhang2025scizor} relies on task progress estimation and transition-level labeling, operating under the assumption that most demonstrations are good. This assumption weakens robustness when non-expert or noisy data dominates.

Overall, while learning-based ranking methods such as DemInf, CUPID, and BED can be effective, they are computationally expensive, difficult to scale, and ill-suited for practical deployment. In contrast, our method is fast, fully offline, and requires no environment interaction, making it well-suited for field deployment and learning policy from large-scale datasets containing many non-expert/sub-optimal demonstrations.

\section{Problem Formulation and Preliminaries}
\subsection{Preliminaries on Imitation Learning: }
We consider offline imitation learning as a Markov Decision Process (MDP) represented by the tuple \((S, A, r, q, s_0, \gamma)\). Here, the reward function \(r\) and the environment dynamics \(q\) are unknown. However, we assume access to a dataset \(D = \{\tau_1, \tau_2, \ldots, \tau_N\}\) comprising of \(N\) trajectories, which include \textbf{both optimal and erroneous demonstrations}. Every trajectory \(\tau_i\) is a sequence of state-action pairs, \(\tau_i = \{(s_1, a_1), (s_2, a_2), \ldots, (s_{T_i}, a_{T_i})\}\), where \(s \in S\) denotes a state, \(a \in A\) denotes an action, and \(T_i\) is the horizon of the trajectory \(\tau_i\). In case of behavior cloning (BC) -- the most prevalent policy learning method in contemporary IL literature -- the objective is to learn a policy \(\pi_\theta : S \rightarrow A\) that maps states to actions by minimizing the standard BC loss function $\mathbb{E}_{(s,a)\sim D}[-log \pi_{\theta}(a|s)] $.
 
\begin{figure}[thpb]
  \centering
  \includegraphics[scale=0.45]{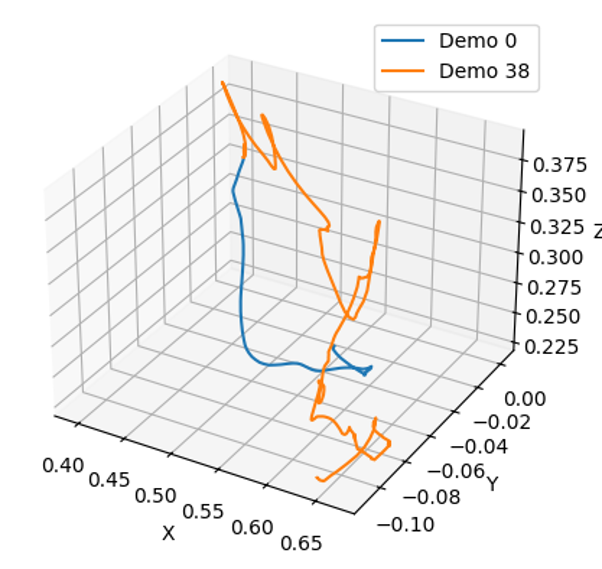}
  \caption{Example of an expert (Demo 0) and a non-expert (Demo 38) trajectory. Expert trajectories are generally smooth and contains less surprised turns compare to a non-expert trajectory.}
  \label{fig: expert}
\end{figure}

\subsection{Expert Vs Non-expert Demonstration}
Expert demonstrations are typically smoother while non-expert demonstrations exhibit motion artifacts including high jerk, temporal inconsistency, abrupt or oscillatory motions, prolonged pauses, and repeated trial-and-error behaviors.  
Figure \ref{fig: expert} shows representative example of expert (an author of this paper) and non-expert (an older adult participant of our user study) trajectories for a task of opening a microwave by a Franka robot arm. Note that we consider adversarial demonstrations as coordinated efforts to fool the learning algorithm while non-expert demonstrations as inadvertent, natural human errors; our work only deals with non-expert demonstrations that are prevalent in field deployment.


\subsection{Frequency-Domain Analysis of a Signal} \label{Fourier}
For an aperiodic time-domain signal $x(t)$ with $T$ samples, the discrete Fourier transform (DFT) 
is computed as \cite{oppenheim1999discrete}:
\begin{equation}
    X(\Omega) = \sum_{t=1}^{T} x(t)\, e^{-j \Omega t}
    \label{dft}
\end{equation}
with frequency $\Omega= \frac{2\pi}{T}k, k=\{0,1,\ldots,T-1\}$.
The power spectral density (PSD) of the signal is proportional to its frequency content:
\begin{equation}
   P(\Omega) \infty \left| X(\Omega) \right|^2  
   \label{psd}
\end{equation}
The noisy (i.e., high-frequency) version of a signal -- e.g. the trajectory of a task as shown in Fig. \ref{fig: expert} -- has higher PSD than its non-noisy (i.e., low-frequency) version. Non-expert trajectories typically contain motion artifacts — such as oscillations, abrupt corrections, and erratic adjustments — that elevate their frequency content relative to expert demonstrations (e.g., Fig. \ref{fig: expert}). That makes PSD comparison between two trajectories of the same task a reliable proxy for demonstration quality. 

\section{PSD of Demonstration Trajectory: An Efficient Metric for Quality Measurement}
We propose a scalar metric 
$W$ derived solely from the PSD of the low-level demonstration trajectory. Despite its simplicity, this metric effectively captures the overall motion quality of a demonstration — higher values indicate increased oscillatory and corrective behavior typical of non-expert users, while lower values correspond to the smoother, more deliberate motions characteristic of expert demonstrations. Notably, $W$ focuses exclusively on trajectory quality, without regard to whether the task was ultimately completed. This design choice is intentional: IL policies trained on mixed-quality datasets benefit not only from expert demonstrations with task success, but also from the recovery strategies and broader state-space coverage present in non-expert demonstrations \cite{huang2025using}, regardless of task outcome. However, trajectories that deviate excessively from a nominal path risk driving a BC policy into irrecoverable states, leading to task failure. The proposed PSD-based metric $W$ directly captures this trade-off. Importantly, $W$ is computed entirely offline, requires no task-specific rewards or environment interaction, and is agnostic to the choice of policy representation.

\subsection{Trajectory Representation}
From each demonstration trajectory $\tau_i \in D$, we extract a time domain signal
\[
x_i(t) \in \mathbb{R}^d, \quad t = 1, \ldots, T_i,
\]
where $x_i(t)$ are kinematic state variables, $d$ is the signal dimensionality, and $T_i$ is the signal length. We specifically used end-effector position in the 3D Cartesian space as $x_i(t)$, making $d=3$. To ensure consistency across demonstrations, we treat each dimension independently and analyze temporal variation along each dimension.
\subsection{Frequency Domain Decomposition of a Trajectory}
Using (\ref{dft}) and (\ref{psd}) we calculate the PSD $P_{i,d}(\Omega)$ for each dimension of $x_i(t)$. We aggregate spectral powers across all dimension $d=\{0,1,2\}$ to obtain a trajectory-level PSD:
\[
P_i(\Omega) = \sum_{d=0}^{2} P_{i,d}(\Omega).
\]
Finally, the total spectral power of the trajectory is calculated as the metric to measure the quality of the demonstration $\tau_i$. 
\[
W_i=\mathrm{PSD}(\tau_i) = \sum_{\Omega} P_i(\Omega).
\]
Fig. \ref{fig:concept} demonstrate this process. The higher spectral power of any finite time domain signal resides in the lower frequencies \cite{oppenheim1999discrete}. However, the abrupt movements in non-expert trajectories add more high frequency components in their DFT and thereby contributing to higher $W$ values (as shown in the 2nd column of Fig \ref{fig:concept}).   


\subsection{Demonstration Ranking and Evaluation}
 Demonstrations $\tau_i$ are ranked froom good to bad by ascending $W_i$ values. 
\[
\tau_i \prec \tau_j \quad \text{if} \quad W_i < W_j .
\]
Based on this ranking, we curate datasets by discarding a fraction $\rho$ of the lowest-ranked demonstrations. Then $D_\rho$ dataset can be used by any policy learning algorithms.

\[
D_\rho = \{\tau_i \in D \mid W_i \le q_{1-\rho}(W)\}.
\]

Where $q_\alpha(W)$ denotes the emperical $\alpha$ quantile of the score distribution. Since larger $W_i$ indicates lower-quality demonstrations, thresholding at $q_{1-\rho}(W)$ discards the highest-scoring fraction $\rho$ of trajectories and retains the remaining $1-\rho$.


\section{Experiments}

We conducted experiments to answer the following two research questions:
\begin{itemize}
    \item \textbf{R1. Demonstration quality}: How effectively does the proposed PSD metric rank demonstrations in a mixed-quality training set relative to competing ranking methods? 
    \item \textbf{R2. Policy Performance}: How does filtering demonstration data with the proposed PSD metric impact downstream imitation learning policy performance?
\end{itemize}
Experiments consider four baselines for comparison: 1) performance without any data curation, 2) performance with curation done by an oracle 3) performance with curation done by two contemporary demo ranking methods: DemInf \cite{hejna2025robot} and CUPID \cite{agia2025cupid} 4) performance with curation done using two  kinematic variables of motion -- jerk and path-length -- suggested in \cite{sakr2025consistency}: 

\subsection{Experimental Setup}
We consider two experimental setups for evaluation: 1)  offline policy learning from a mixed quality data set and 2) fine-tuning a pre-trained generalist model, $\pi$0.5 \cite{intelligence2025pi_}, with non-expert data.

\subsubsection{Offline policy learning from mixed-quality data}
We considered two benchmark IL datasets manipulation tasks -- Robomimic-MH \cite{robomimic2021}, and Layman2 \cite{sojib2024self}. Each dataset contains demonstrations for two simulation tasks: \textbf{Can} (the robot must grasp a cylindrical can and place it at a designated target location) and \textbf{Square} (the robot must pick up a square nut and place it onto a square peg). \\
$\bullet\;\;${\textbf{Robomimic-MH:}} This is a commonly used IL benchmark consisting 300 demonstrations (270 for training and 30 for validation) for each of the two tasks, collected from six human users, annotated with three expertise levels: \textit{better(label 3)}, \textit{okay (label 2)}, and \textit{worse (label 1)}. Demonstrations are recorded in simulation using standard robomimic environments.\\
$\bullet\;\;${\textbf{Layman2:}} This dataset contains 300 demonstrations for each of the two tasks by 15 non-expert users with diverse skills. The dataset mirrors the task definitions and annotation structure of Robomimic-MH. Demonstrations in this dataset are longer in duration, exhibit more trajectory variations and trial--error behaviors, typically exhibited by lay users.\\
To investigate \textbf{R1}, we measure the average quality of the remaining demonstrations after
removing the bottom 50\% according to various baseline methods. Demonstrations are ranked offline using different metrics, the bottom fraction of demonstrations is removed, and IL policies are trained on the filtered datasets. Additionally, we evaluate the time required for ranking by the proposed method and baselines as show in the table \ref{tab:training_time}. 

For all experimental conditions, policies are trained using the same diffusion-based imitation learning architecture with identical optimization and training hyperparameters. Performance is evaluated over three random seeds and reported as the average success rate across evaluation episodes.
    
    
    
    
\begin{figure*}
\centering
\includegraphics[width=7.1in]{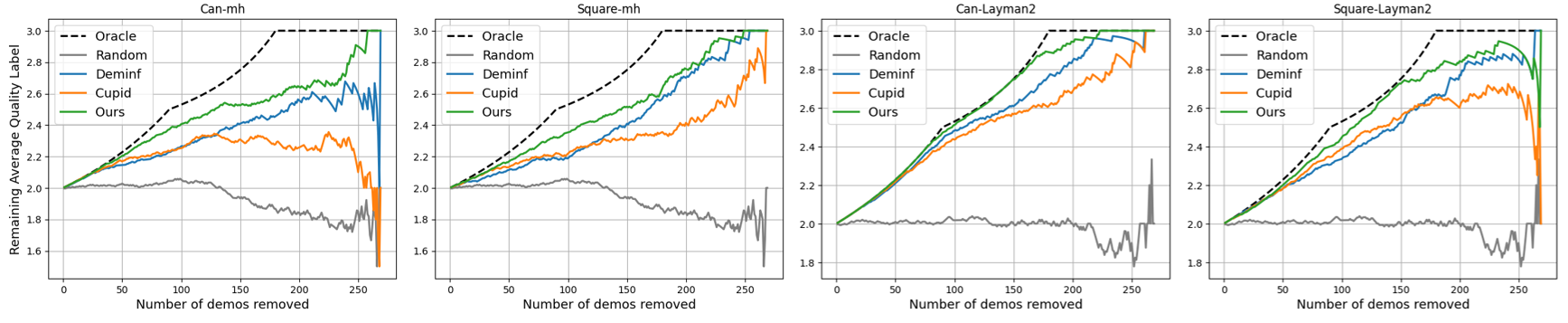}
\caption{Remaining average demonstration quality is plotted against the number of demonstrations removed for Can and Square tasks under MH and Layman2 datasets. The dashed black line denotes an oracle ordering, while gray corresponds to random removal. Among learned methods, Ours maintains higher remaining quality as demonstrations are progressively removed and more closely follows the oracle compared to DemInf and Cupid across tasks and datasets.}
\label{fig:remainingq}
\end{figure*}
\subsubsection{Fine tuning $\pi$0.5}
\begin{figure}[thpb]
  \centering
  \includegraphics[scale=0.15]{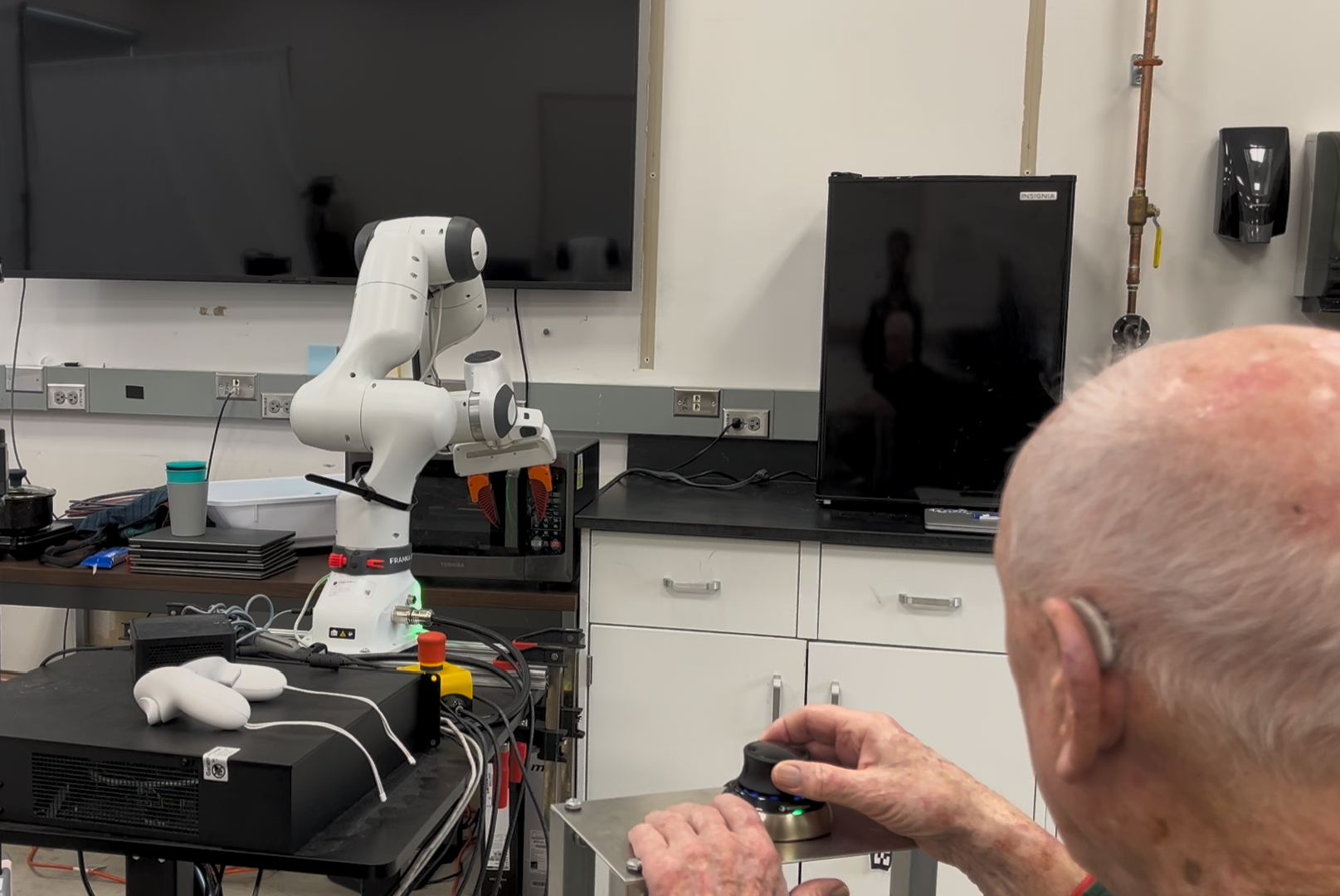}
  \caption{An older adult participant from our study teleoperating a Franka robot using a SpaceMouse to open a microwave in a simulated home setting.}
  \label{fig: participant}
\end{figure}

Due to the nature of this setup, we only investigated \textbf{R2}. We conducted an IRB‑approved user study with five older adults (ages 79, 80, 86, 84 and one undisclosed) to simulate a realistic home‑deployment environment. The robot is equipped with $\pi$0.5, a generalist policy, which needs to be fine-tuned by the participants to be able to open a microwave in the kitchen, as shown in Fig. \ref{fig: participant}. We asked the participants to give demonstrations using two interfaces -- a space mouse and a meta quest VR. Participants were given 30 minutes to play with both interfaces to gain familiarity and provided data only when they felt comfortable handling the robot and interfaces. We stopped data collection when the participant filled fatigued or felt uncomfortable, to emulate a realistic deployment setting where participants are unlikely to spend extended time to generate many demonstrations for the robot. The five participants generated 20,20,17,15, and 20 demonstrations. We fine tuned $\pi$0.5 with a batch size of 16 and for 10000 steps.

\section{Results}

\begin{table}[t]
\centering
\caption{Average quality of the remaining demonstrations after filtering out 50\% of the data (higher is better).}
\label{tab:demo_quality}
\setlength{\tabcolsep}{5pt}
\begin{tabular}{| l | l | c | c | c | c | c |}
\hline
Dataset & Task & Random & Oracle  & DemInf & CuPID & Ours\\
\hline
\multirow{2}{*}{Robomimic} 
 & Can    & 1.99 & 2.67 & 2.37 & 2.32 & 2.53 \\
\cline{2-7}
 & Square & 1.99 & 2.67 & 2.34 & 2.30 & 2.47 \\
\hline
\multirow{2}{*}{Layman2} 
 & Can    & 1.99 & 2.67 & 2.57 & 2.53 & 2.67 \\
\cline{2-7}
 & Square & 1.99 & 2.67 & 2.48 & 2.50 & 2.64 \\
\hline
\end{tabular}
\end{table}

\begin{table*}[t]
\centering
\caption{Success rate across 3 seeds on $50\%$ filtered data.
\textbf{State} and \textbf{Image} denote different experiment modalities.}
\label{tab:demo_quality_combined}
\resizebox{\textwidth}{!}{
\begin{tabular}{| l | l | l | c | c | c | c | c |}
\hline
\textbf{Experiment} & \textbf{Dataset} & \textbf{Task} 
& \textbf{Unfiltered} & \textbf{Oracle} & \textbf{DemInf} & \textbf{Cupid} & \textbf{Ours} \\
\hline

\multirow{4}{*}{\textbf{State}}
& \multirow{2}{*}{Robomimic}
& Can
& $0.91\pm0.01$ & $\boldsymbol{1.00\pm0.00}$ & $\boldsymbol{1.00\pm0.00}$ & $0.91\pm0.04$ & $\boldsymbol{1.00\pm0.00}$ \\
\cline{3-8}
& 
& Square
& $0.87\pm0.03$ & $0.85\pm0.02$ & $0.89\pm0.02$ & $0.77\pm0.04$ & $\boldsymbol{0.90\pm0.02}$ \\
\cline{2-8}

& \multirow{2}{*}{Layman2}
& Can
& $0.31\pm0.01$ & $0.71\pm0.02$ & $0.78\pm0.03$ & $0.21\pm0.04$ & $\boldsymbol{0.86\pm0.00}$ \\
\cline{3-8}
& 
& Square
& $0.44\pm0.05$ & $0.73\pm0.04$ & $0.66\pm0.04$ & $0.27\pm0.01$ & $\boldsymbol{0.79\pm0.01}$ \\
\hline

\multirow{4}{*}{\textbf{Image}}
& \multirow{2}{*}{Robomimic}
& Can
& $0.97\pm0.01$ & $0.98\pm0.02$ & $\boldsymbol{1.00\pm0.00}$ & $0.96\pm0.02$ & $0.99\pm0.01$ \\
\cline{3-8}
& 
& Square
& $0.87\pm0.07$ & $\boldsymbol{0.90\pm0.02}$ & $0.87\pm0.02$ & $0.87\pm0.02$ & $0.85\pm0.05$ \\
\cline{2-8}

& \multirow{2}{*}{Layman2}
& Can
& $0.05\pm0.01$ & $0.27\pm0.08$ & $0.21\pm0.08$ & $0.19\pm0.02$ & $\boldsymbol{0.21\pm0.04}$ \\
\cline{3-8}
& 
& Square
& $0.21\pm0.02$ & $0.29\pm0.06$ & $0.28\pm0.07$ & $0.08\pm0.03$ & $\boldsymbol{0.34\pm0.04}$ \\
\hline
& & Mean
& $0.58\pm0.03$ & $0.72\pm0.03$ & $0.71\pm0.03$ & $0.53\pm0.03$ & $\boldsymbol{0.74\pm0.02}$\\
\hline
\end{tabular}
}
\end{table*}

\subsection{Offline policy learning from mixed-quality data} 



Table \ref{tab:demo_quality} shows the filtering out using our method can effectively remove bad demonstrations compared to others as the average quality of the remaining demonstrations after removing 50\% data is higher. Ranking using W consistently retains higher-quality demonstrations compared to all other baselines, and closely follows the oracle ordering. This trend is particularly pronounced on Layman2, where non-expert artifacts dominate and the other two learning-based ranking methods \cite{hejna2025robot,agia2025cupid} exhibit weaker correlation with ground-truth quality.

Figure \ref{fig:remainingq} further illustrates ranking behavior as demonstrations are progressively removed. The proposed method maintains higher remaining quality across removal thresholds and more closely tracks the oracle curve than DemInf \cite{hejna2025robot} and Cupid \cite{agia2025cupid}.

We next evaluate whether improved ranking translates to downstream policy performance (\textbf{R2}). Table \ref{tab:demo_quality_combined} reports success rates after training on datasets filtered using each method.

Across both Robomimic and Layman2, policies trained on demonstrations filtered by our the proposed metric consistently outperform those trained on mixed-quality data and are competitive with oracle selection. Notably, our method achieves the largest gains on Layman 2, highlighting its effectiveness in realistic non-expert settings where control inefficiencies are prevalent.


\subsubsection{Comparison of W with other Kinematic proxies}
\begin{figure}[thpb]
  \centering
  \includegraphics[scale=0.33]{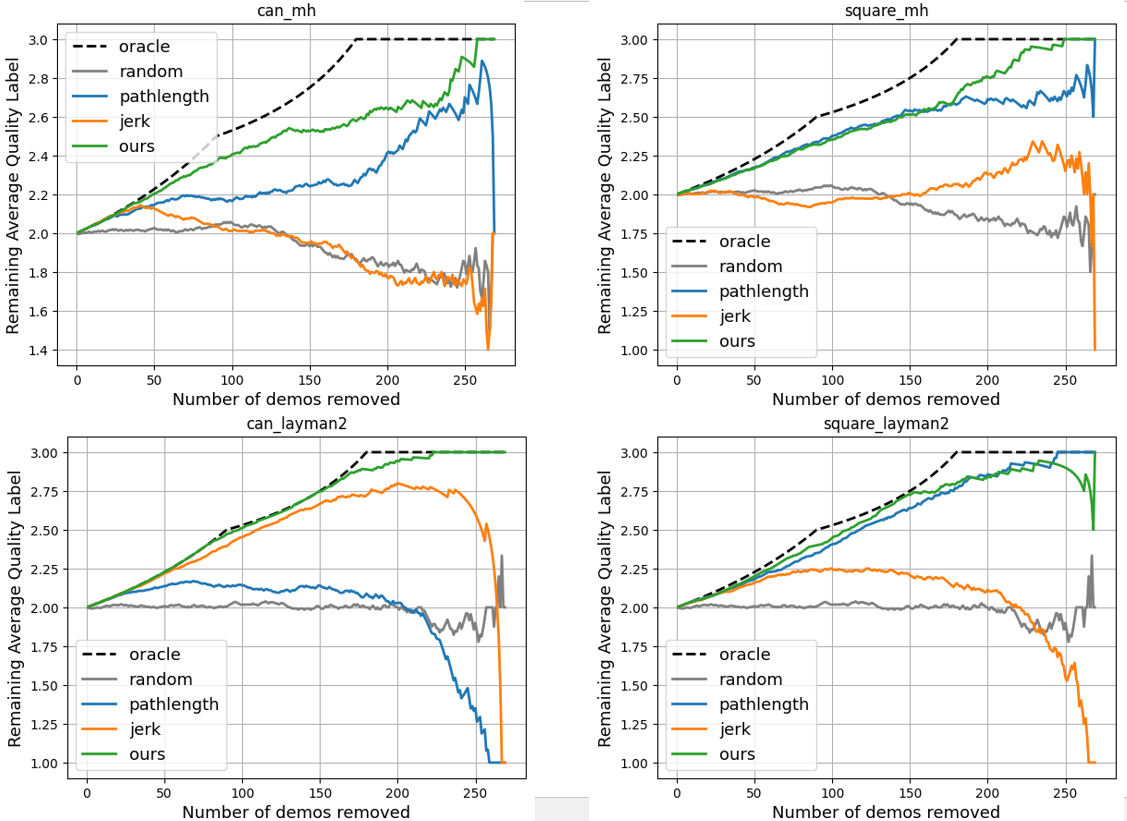}
  \caption{Our method generalize simple kinematics methods such as jerk and path length.}
  \label{fig: fft_jerk_pl}
\end{figure}

To assess whether the proposed W provides any benefit beyond analyzing the trajectory path length and mean jerk norm -- quantities that have been shown to correlate with demonstration consistency in prior work~\cite{sakr2025consistency} -- we compare against these proxies. Across all tasks and datasets, the proposed W consistently outperforms these baselines in terms of preserving demonstration quality. Figure \ref{fig: fft_jerk_pl} illustrates this distinction. While trajectory path length and jerk magnitude can correlate with demonstration quality in isolated cases, their behavior is inconsistent across tasks and datasets. In particular, path length is highly sensitive to variations in initial object poses and goal distances, which naturally differ across demonstrations even when execution quality is comparable. In contrast, the metric W provides a more stable and task-agnostic signal by directly capturing corrective and oscillatory motion patterns that arise from inefficient control.

\subsubsection{Computational Efficiency of Quality Measurement}
A key advantage of the proposed spectral metric is its computational efficiency. Unlike learning-based ranking methods such as DemInf and Cupid, which require expensive nearest-neighbor searches or policy rollouts to estimate demonstration utility, our method operates directly on raw trajectory time series using a single discrete Fourier transform per demonstration.

\begin{table}[t]
\centering
\caption{Approximate computation time required to generate demonstration rankings for each method (including all training and final ranking generation steps).}
\label{tab:training_time}
\begin{tabular}{| l | l | c | c | c |}
\hline
\textbf{Dataset} & \textbf{Task} & \textbf{DemInf} & \textbf{Cupid} & \textbf{Ours} \\
\hline
\multirow{2}{*}{Robomimic}
& Can     & 300 minutes & 360 minutes & 10s\\
\cline{2-5}
& Square  & 300 minutes & 360 minutes & 10s\\
\hline
\multirow{2}{*}{Layman v2}
& Can     & 300 minutes & 360 minutes & 10s\\
\cline{2-5}
& Square  & 300 minutes & 360 minutes & 10s\\
\hline
\end{tabular}
\end{table}
 
We compare the wall-clock time required to compute quality scores for each method under identical hardware and dataset conditions. As shown in Table \ref{tab:training_time}, our approach is orders of magnitude faster than DemInf and Cupid, enabling scalable demonstration curation for large datasets and real-world robotic systems where rapid offline processing is essential.


\subsection{Fine tuning $\pi$0.5}

Since real-world deployment assumes non-expert users will teach robots new tasks—and collecting demonstrations is costly—we adopt a state-of-the-art generalist policy and fine-tune it on newly collected user data using Franka robot on the microwave openning task. For each participant, we fine-tuned $\pi 0.5$ using LoRA \cite{hu2022lora} with a batch size of 16 for 10,000 gradient steps on an RTX 5090 consumer GPU. We evaluated two training settings per user: (1) the full dataset (all demonstrations) and (2) a filtered subset retaining 60\% of demonstrations (removing 40\%). In both cases, we performed 20 rollouts and report the average success rate. As shown in Table~\ref{tab:user_filtered}, training on the filtered dataset consistently improves policy performance.


\begin{table}[htbp]
\centering
\caption{Average success rate of $\pi0.5$ on the microwave opening task using non-expert demonstrations. DemInf and Ours are curation methods retaining 60\% of demonstrations, while Unfiltered uses the full dataset.}
\label{tab:user_filtered}
\begin{tabular}{|c|c|c|c|}
\hline
\textbf{User} & \textbf{Unfiltered} & \textbf{DemInf} & \textbf{Ours} \\
\hline
Participant 1 & 0.15 & 0.50 & \textbf{0.55}  \\
\hline
Participant 2 & 0.15 & 0.25 & \textbf{0.50} \\
\hline
Participant 3 & 0.00 & 0.10 & \textbf{0.30} \\
\hline
Participant 4 & 0.35 & 0.15 & \textbf{0.40} \\
\hline
Participant 5 & 0.10 & \textbf{0.35} & 0.15 \\
\hline
\textbf{Mean} & 0.15 & 0.27 & \textbf{0.38} \\
\hline
\end{tabular}
\end{table}


\subsection{Discussion} 

The proposed \textit{PSD} metric targets control inefficiency—manifested as high-frequency oscillations, jitter, and corrective motion—rather than explicitly encoding semantic task success. While this design choice allows for the possibility of smooth but unsuccessful demonstrations, our experiments indicate that in realistic non-expert datasets such as \textit{Layman v2}, the primary source of performance degradation in offline imitation learning arises from erratic, trial-and-error behavior that introduces substantial spectral energy. In these regimes, \textit{PSD} effectively filters demonstrations that disproportionately harm policy learning. Compared to simple kinematic heuristics such as path length or jerk magnitude, \textit{PSD} captures finer-grained temporal characteristics of control by emphasizing the rate and consistency of adjustments rather than absolute trajectory length, enabling the metric to remain largely task-agnostic. Finally, because \textit{PSD} requires only a single discrete Fourier transform per trajectory and no policy rollouts or environment interaction, it is orders of magnitude more computationally efficient than learning-based ranking methods, making it a practical and scalable tool for curating large-scale robotic demonstration datasets.

\section{CONCLUSIONS}

We presented a simple and scalable approach for automatically curating human demonstration datasets using frequency-domain analysis of motion trajectories. By ranking demonstrations based on spectral power, our method identifies control inefficiencies commonly exhibited by non-expert users and filters low-quality data prior to policy learning.

Despite its simplicity, the proposed metric is fully offline, computationally efficient, and requires no task-specific tuning or policy rollouts. Extensive experiments across simulated benchmarks, a newly collected non-expert dataset, and real-robot manipulation tasks demonstrate that spectral ranking consistently improves imitation learning performance and compares favorably to contemporary ranking methods.

While the proposed metric does not aim to capture all aspects of demonstration optimality, our results indicate that frequency-domain signatures of control inefficiency provide a strong and practical proxy for demonstration quality in motion-dominated manipulation tasks. We believe this work offers a lightweight yet effective tool for scaling demonstration-driven robot learning to realistic, mixed-quality data collection settings.

\addtolength{\textheight}{-12cm}   











\newpage
\bibliographystyle{IEEEtran}
\bibliography{references.bib}

\appendix

\section{Additional Experimental Details}
Training details: To account for differences in demonstration length between datasets, we adjust the rollout horizon for Layman v2 while keeping all other settings fixed. Specifically, we use 1200 timesteps for the \textit{can} task and 1400 timesteps for the \textit{square} task on Layman v2, compared to 500 and 800 timesteps, respectively, for Robomimic-MH. This adjustment reflects the longer and more variable trajectories in Layman v2, which arise from increased trial–error behavior by non-expert users. In all cases, timestep limits are chosen based on the maximum trajectory lengths observed in each dataset to ensure fair and consistent evaluation.

\end{document}